\documentclass{article}
\usepackage{graphicx} 
\usepackage[final]{corl_2024} 
\usepackage{hyperref}

\hypersetup{
    citecolor=green,
}
\usepackage{booktabs}
\usepackage{multirow}

\title{A Dual Process VLA: \\Efficient Robotic Manipulation Leveraging VLM}

\author{
  ByungOk Han, Jaehong Kim, and Jinhyeok Jang  \thanks{Corresponding author.}
\\
  ETRI\\
  Republic of Korea\\
  \texttt{\{byungok.han, jhkim504, jjh6297\}@etri.re.kr}
}

\begin{document}
\maketitle


\begin{abstract}
Vision-Language-Action (VLA) models are receiving increasing attention for their ability to enable robots to perform complex tasks by integrating visual context with linguistic commands. However, achieving efficient real-time performance remains challenging due to the high computational demands of existing models. To overcome this, we propose Dual Process VLA (DP-VLA), a hierarchical framework inspired by dual-process theory. DP-VLA utilizes a Large System 2 Model (L-Sys2) for complex reasoning and decision-making, while a Small System 1 Model (S-Sys1) handles real-time motor control and sensory processing. By leveraging Vision-Language Models  (VLMs), the L-Sys2 operates at low frequencies, reducing computational overhead, while the S-Sys1 ensures fast and accurate task execution. Experimental results on the RoboCasa dataset demonstrate that DP-VLA achieves faster inference and higher task success rates, providing a scalable solution for advanced robotic applications.
\end{abstract}





\section{Introduction}
Vision-Language-Action (VLA) models are designed to enable robots to generate actions based on a user's task instruction by following three key steps: 
(1) interpreting the task instruction, 
(2) analyzing the current visual information in relation to the task, 
and (3) predicting the necessary actions for execution. 
By combining vision and language inputs, VLA models allow robots to perform complex tasks using both visual context and linguistic commands. 
Recently, Large Language Models (LLMs) \cite{touvron2023llama, vicuna2023, jiang2023mistral} and Vision-Language Models (VLMs) \cite{openai2024gpt4technicalreport, liu2023llava, alayrac2022flamingo} have reported high capabilities to general understanding. VLA models have leveraged VLMs to enhance a robot's perception capabilities, showing promising results in their ability to interpret and execute complex tasks.
By this, recent VLA have demonstrated accurate action generation across various tasks, utilizing diverse robot hardware in real-world environments such as RT-2 \cite{brohan2023rt}, RoboFlamingo \cite{li2023vision}, OpenVLA \cite{kim2024openvla}, LLaRA \cite{li2024llara}, and LLARVA \cite{niu2024llarva}.




One of the key factors enabling this advancement is the availability of large robot datasets, which are essential for training VLAs with VLMs that require vast amounts of data to generate appropriate actions effectively.
The Open-X-Embodiment (OXE) dataset \cite{oxe_dataset}, which features data collected from various robot platforms by multiple research groups in real-world environments and formatted in a standardized manner, has been fully released, making a significant contribution to advancements in the field.
The OXE dataset was created using data from 22 different robots, collected in collaboration with 21 institutions, and demonstrates 527 distinct skills.
Similarly, the DROID dataset \cite{khazatsky2024droid}, using the Franka Emika Panda robot, provides extensive robot trajectories across a range of tasks and environments. 
The DROID dataset consists of 76,000 demonstration trajectories, or 350 hours of interaction data, collected across 564 scenes and 86 tasks.
In addition, with advancements in rendering technology, simulation-based robot datasets, such as Calvin \cite{mees2022calvin} and RoboCasa \cite{robocasa2024}, have also been made available to the research community.

Despite recent advancements in VLA models based on large robot datasets, VLA methods using VLMs still encounter significant challenges in real-time applications due to high computational demands.
This leads to slow inference, causing unnatural and discontinuous motions in robots that require fast action prediction.
For example, RT-2 \cite{brohan2023rt} demonstrated that their 55B model operates at 1 to 3 Hz, while the 5B model runs at around 5 Hz under experimental conditions. 
Similarly, OpenVLA \cite{kim2024openvla} achieves a processing speed of 6 Hz on a commercially available GPU. 
On the other hand, BC Transformer \cite{mandlekar2022matters} and ALOHA \cite{fu2024mobile} operate at high speeds (around 50Hz), but they lack general reasoning capabilities. As a result, their performance declines in unseen environments.
Moreover, VLA models still require fine-tuning to adapt to specific environments and robot hardware, which imposes a computational burden, consuming significant GPU memory and time during training.

In this paper, we propose Dual Process VLA (DP-VLA), a hierarchical framework inspired by \textbf{Dual-process Theory} \cite{kahneman2011thinking, evans2008dual} to enhance robotic action prediction by separating tasks into two subsystems: Large System 2 Model (L-Sys2) and Small System 1 Model (S-Sys1). 
The L-Sys2, acting as the high-level decision-maker, processes complex reasoning and task planning based on visual and language inputs, while the S-Sys1 focuses on real-time motor control using diverse continuous sensory inputs.
Our contributions are as follows: (1) \textbf{Efficient robot manipulation}—the framework ensures precise and responsive manipulation with \textit{improvements in both speed and accuracy}; 
(2) \textbf{Scalable design}—it allows seamless upgrades to advanced VLMs \textit{without requiring modifications} to the overall system; and
(3) \textbf{Experimental validation}—through our experiments, we demonstrate the effectiveness of our DP-VLA in the RoboCasa simulation environment, achieving \textit{superior performance} compared to previous VLA approaches.

\section{Motivation}
\label{sec:motivation}
\vspace{-10pt}
Our solution draws inspiration from human cognitive psychology. According to the literature about Dual-process Theory \cite{kahneman2011thinking, evans2008dual}, there are two distinct modes of thinking: System 1 and System 2.

\textbf{System 1}: This mode is fast, automatic, and intuitive. It relies on heuristics (mental shortcuts) to make quick decisions with little conscious effort. While it is prone to biases, System 1 is highly efficient for routine tasks and rapid responses.

\textbf{System 2}: This mode represents a slower, more deliberate, and analytical mode of thinking. It involves conscious thought, careful evaluation, and logical reasoning, making it suitable for complex decision-making, problem-solving, and overriding instinctive responses generated by System 1.  

System 1 is primarily associated with more primitive and automatic brain structures, such as the limbic system, whereas System 2 is linked to the prefrontal cortex, which is responsible for higher-order reasoning and conscious control. This concept has often been adapted in AI research to enhance efficiency or accelerate processing. For instance, Qi et al. \cite{qi2024interactive} implemented a separation where Vision Transformer (ViT) acted as System 1 and a VLM served as System 2, aiming to improve continual learning. Similarly, Yoshua Bengio proposed System 2-like processes in AI, suggesting a mechanism to distinguish rapid, unconscious (System 1) processing from more deliberative (System 2) approaches. This separation enables AI systems to handle frequent and routine tasks with efficiency, while reserving complex reasoning for specialized processing \cite{bengio2017consciousness}.

Leveraging the concept of dual process theory, we designed our approach to divide the overall function into two sub-functions: simple tasks, such as action generation, and complex tasks, like reasoning. Specifically, we utilize a VLM to serve as System 2, which is processed at a low frequency to handle reasoning and complex decision-making. In parallel, we employ a small robot policy to function as the short-term planning and motor control module, adapting the determined intention to the diverse sensory inputs.
This modular design not only reduces redundant computations but also enables smoother and more natural robotic actions, thereby addressing the limitations of previous VLA approaches, which often suffer from slow processing times and discontinuous actions. By scheduling the intention module (System 2) to run less frequently over multiple time steps, we achieve improved computational efficiency without sacrificing performance. This hierarchical separation ensures that the simple, frequent tasks are managed efficiently, while the more complex and infrequent operations receive the necessary computational focus.


\section{Method: Dual Process VLA}
\label{sec:method}
\begin{figure}[!t]
	\centering
    \begin{minipage}{\linewidth}
	    \centering
	    {\includegraphics[width=1.\textwidth]{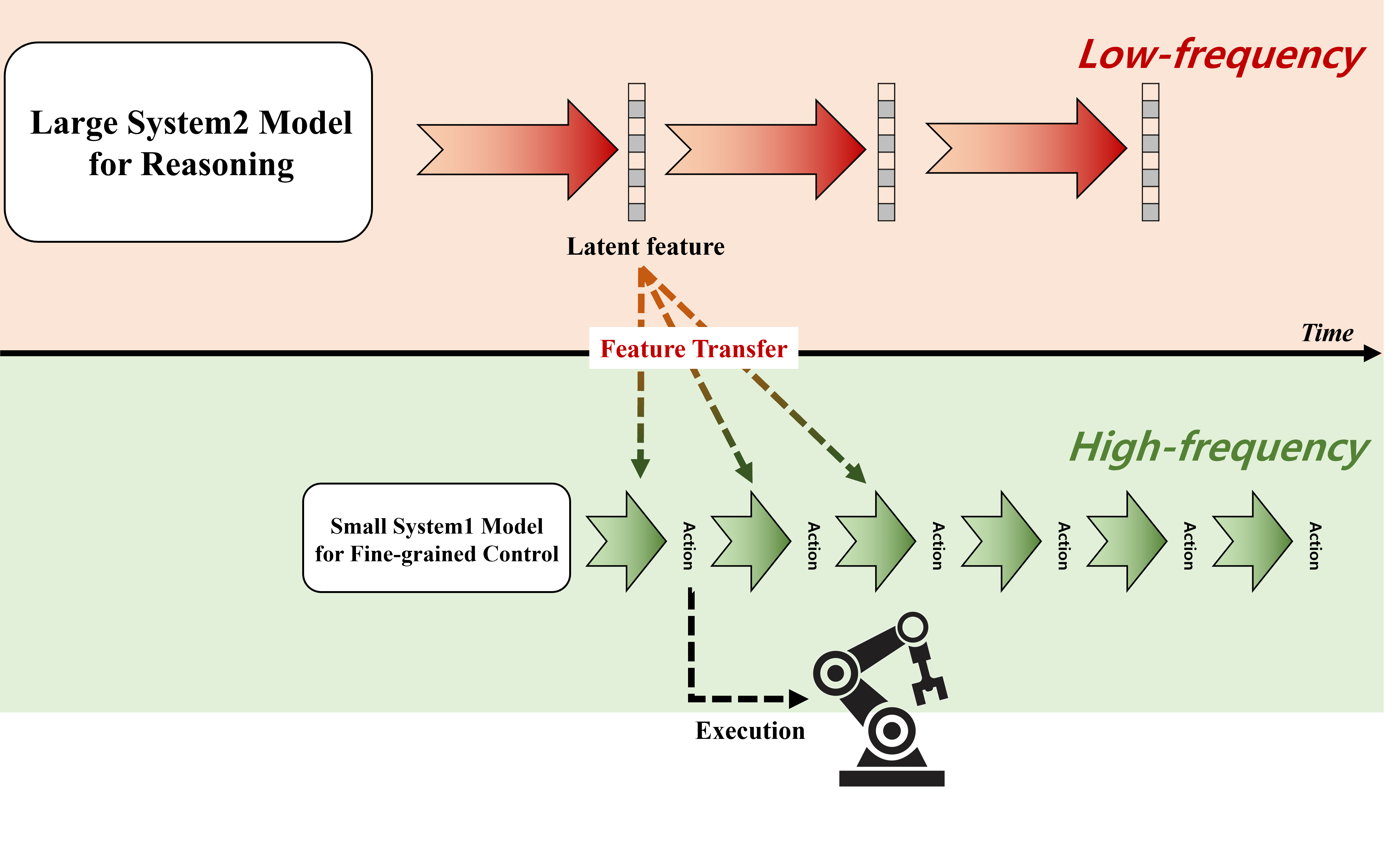}}  
    \end{minipage}%
\vspace{-20pt}
\caption{\small Our Dual Process framework integrates the concepts of System 2 and System 1. The Large System 2 Model (L-Sys2) extracts latent features that encode both reasoning information related to user instructions and environmental context. These high-level representations guide the Small System 1 Model (S-Sys1) in generating fine-grained actions in real-time, leveraging various observations and the robot's states. }
\label{fig:overall_flow}
\end{figure}

From the perspective of robotic actions, we have structured the models into System 1 and System 2 to improve the prediction of both logical and immediate actions as illustrated in Figure \ref{fig:overall_flow}. 
The L-Sys2 is responsible for more complex, logical decision-making processes, while the S-Sys1 is designed to handle immediate responses. 
We designed separate L-Sys2 and S-Sys1 models for accurate action inference for several key reasons.
\begin{itemize}
   \item[(1)] \textbf{Model size and real-time inference}: Modern VLA models, including VLMs ranging from 7B to over 50B parameters, are too large for real-time inference. These models require substantial computational resources, making on-device inference impractical.
   \item[(2)] \textbf{Data requirements}: Large VLA models require vast amounts of data—such as trajectories from diverse environments and robots—to achieve precise action predictions. Training these models on extensive data is challenging and often demands expensive hardware, particularly GPUs, which increases complexity and resource consumption.
   \item[(3)] \textbf{Fine control}: The separation of L-Sys2 and S-Sys1 allows for flexible, precise control. The S-Sys1 can process diverse observations and robot states, incorporating detailed inputs such as gripper images and multiple camera views, enabling accurate control. This approach reduces the computational burden, as VLMs are sensitive to the number of input tokens, which can grow significantly with an increasing number of images or larger image sizes, making it resource-intensive to process.
   \item[(4)] \textbf{Model adaptability}: A major advantage of this separation is the rapid progress in VLMs. By isolating the L-Sys2, we can easily upgrade to newer models, immediately leveraging their advanced capabilities without altering the rest of the system.

\end{itemize}

Based on these reasons, we propsoe a flexible framework that leverages the advanced environment understanding capabilities of rapidly evolving VLMs, utilizing pre-trained models. 
This framework also incorporates robot policies that are relatively easy to train, ensuring fast and accurate action inference. 
This approach effectively balances the strengths of pre-trained VLMs with the efficiency of a simplified robot policy, providing a robust solution for real-time action prediction.

\subsection{Large System 2 Model}
\label{sec:method_lsm}

The L-Sys2 receives both the visual input containing a comprehensive view of the environment and the user's language instructions. 
It utilizes a large model, including an LLM, to generate latent features that contain logical and analytical information regarding the movements that should be made within the environment.
The L-Sys2 is designed to operate when the environment changes completely, requiring the robot to adapt. 
In this paper, we apply it specifically when the instructions given to the robot are newly updated (see Figure \ref{fig:overall_flow}).
We assume that the visual information remains largely consistent from the receipt of new instructions to the completion of their execution.
The L-Sys2 can be instantiated as a VLM, such as LLaVA \cite{liu2023llava}, GPT-4v \cite{openai2024gpt4technicalreport}, or CLIP \cite{pmlr-v139-radford21a}, for interpreting images and text, or as a VLA model, such as OpenVLA \cite{kim2024openvla}, Octo \cite{octo_2023}, or RT-2 \cite{brohan2023rt}, for extracting actions.
In other words, this can be achieved by interpreting visual and textual information to extract latent features.
In our framework, a VLM or a VLA model is represented by a function $f_{\ell}$, which takes as input both visual information $v_t$ and language instruction $\ell_i$. 
For latent feature extraction as visual context information, we simply use the first image $v_0$ in a sequence of $n$ continuous images $\mathbf{v}$ from one arbitrary robot camera view, chosen by the user when the task changes as follows:  
\begin{equation}
    \mathbf{v} = \{v_0, v_1, ... , v_{n-1}\}
\end{equation}

The latent feature $\mathbf{z_i}$ is then computed as:

\begin{equation}
    \mathbf{z_i} = f_{\ell}(v_0, \ell_i)
\end{equation}

\subsection{Small System 1 Model}
\label{sec:method_ssm}
The L-Sys2 occasionally provides a latent feature when there are significant changes in the environment or the user's commands, which acts as a language condition for the S-Sys1.
In contrast, the S-Sys1 continuously receives diverse sensory inputs from the environment and processes them in a real-time manner to predict the robot's actions.
Once the instruction is updated, the L-Sys2 generates a latent feature, which the S-Sys1, a robot policy, encodes along with the robot's observations and states.
Specifically, the inputs to the S-Sys1 can be configured in various ways using the robot's observations and states, independently of the L-Sys2’s inputs.
As long as the instruction remains unchanged, the S-Sys1 encodes the robot's dynamic inputs together with the fixed latent feature $\mathbf{z_i}$ generated by the L-Sys2 to perform action inference at time $t$ as:

\begin{equation}
    \mathbf{a}_t = f_s(\mathbf{o}_t, \mathbf{s}_t, \mathbf{z_i} ), \quad\quad t=0, 1, ..., (n-1)
\end{equation}
where $\mathbf{a}_t$ represents the robot's actions at time $t$ as determined by a robot policy $f_s$, $\mathbf{o}_t$ represents robot's observations, and $\mathbf{s}_t$ represents robot's states.


\section{Experiments}
\label{sec:result}
In this section, we demonstrate the effectiveness of the proposed method by employing OpenVLA \cite{kim2024openvla} for L-Sys2 and BC-Transformer \cite{mandlekar2022matters} for S-Sys1. 
We describe the experimental setup, followed by a quantitative comparison between our DP-VLA method and existing techniques, including L-Sys2-only and S-Sys1-only action prediction. 
Furthermore, we conduct experiments based on the latent features of the L-Sys2 to identify which features contribute most effectively to performance.

\subsection{Experiment Details}
We conducted experiments on the RoboCasa dataset \cite{robocasa2024}, utilizing OpenVLA for L-Sys2 and BC-Transformer for S-Sys1 to learn standard seven-dimensional robot actions (six for end-effector and one for gripper command).
OpenVLA, initialized with publicly available pretrained weights, was fine-tuned on the RoboCasa dataset to adapt to the target environment and tasks. 
In contrast, BC-Transformer was trained from scratch, with the language encoder utilizing pretrained CLIP \cite{pmlr-v139-radford21a}.
Our DP-VLA approach utilized both pre-trained and fine-tuned versions of OpenVLA for L-Sys2. 
For S-Sys1, BC-Transformer was trained from scratch using the latent features extracted from both versions of the L-Sys2, which served as language-conditioning signals.

\begin{figure}[h]
	\centering
    \begin{minipage}{\linewidth}
	    \centering
	    {\includegraphics[width=1.\textwidth]{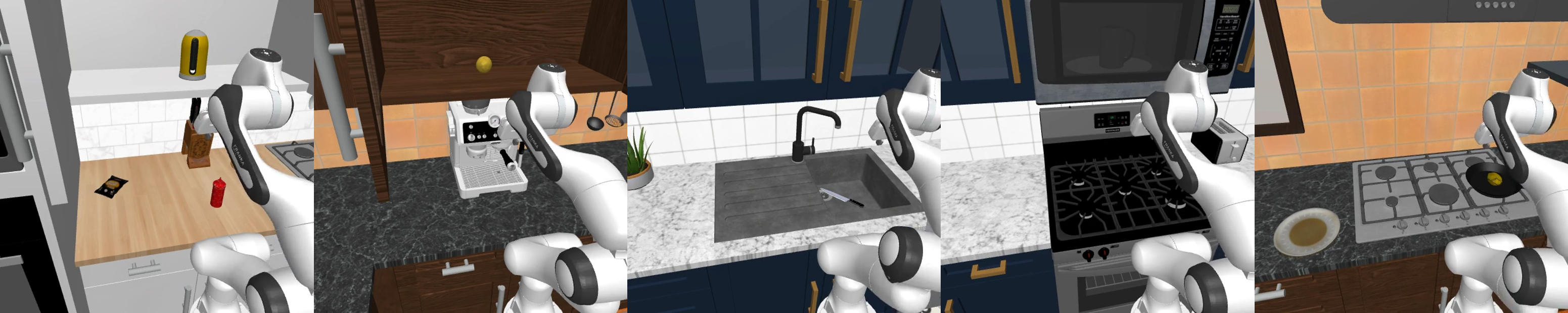}}  
    \end{minipage}%
\caption{\small RoboCasa simulation environment. A Franka Emika Panda robot was used for training and evaluation across diverse kitchen settings.}\label{fig:robocasa}
\end{figure}

\textbf{RoboCasa dataset \cite{robocasa2024}:} 
the dataset consists of robot manipulation data extracted from the MuJoCo simulation environment by using a Franka Emika Panda, with a primary focus on kitchen scenarios as shown in Figure \ref{fig:robocasa}. 
For our experiments, we used the Generated-3000 dataset provided by RoboCasa, containing 72,000 episodes automatically generated via the MimicGen \cite{mandlekar2023mimicgen} algorithm based on human demonstrations. 
The tasks we employed comprise 24 atomic tasks, including actions such as pick-and-place (PnP), open, and close, which are frequently performed in kitchen environments as described in Table \ref{tab:comp_other}.

%

\textbf{Hyper-parameters:}
since OpenVLA can only process a single image as input, we set its observation image to the ``robot0\_agentview\_left\_image" type provided by RoboCasa.
In contrast, BC-Transformer was configured to utilize three types of images: ``robot0\_agentview\_left\_image," ``robot0\_agentview\_right\_image," and ``robot0\_eye\_in\_hand\_image." 
Additionally, the robot’s state information was incorporated, including ``robot0\_base\_to\_eef\_pos," ``robot0\_base\_to\_eef\_quat," ``robot0\_base\_pos," ``robot0\_base\_quat," and ``robot0\_gripper\_qpos" for BC Transformer.
We utilized ResNet18 to encode each image observation, where the size of all images is 128 by 128.
Latent features were extracted as 4,096-dimensional vectors and further encoded through an MLP layer. 
L1 and L2 loss functions were utilized during training with a batch size of 128, and all experiments were conducted a RTX 6000 Ada GPU.

\textbf{Evaluation environment:}
We followed RoboCasa's evaluation protocol by rolling out the trained models. 
The evaluation focuses on testing the models in environments with unseen kitchen styles and on unseen object instances.

\subsection{Comparisons with Other Methods}
To evaluate our approach by comparing it to prior methods, we assessed two key aspects in the RoboCasa simulation environment: task success rate and inference speed, both of which are crucial for practical robot control. This analysis allows us to assess the practical usefulness of our approach.

\subsubsection{Task Success Rate}

As an initial validation, we compared DP-VLA with prior works, such as fine-tuned OpenVLA and BC-Transformer (BC-xfmr), in terms of task success rate and inference time under the RoboCasa evaluation protocol. 
We trained our DP-VLA on the Generated-3000 dataset with a batch size of 128 for 350,000 iterations, using latent features from the OpenVLA-pt model (pre-trained).
As shown in Table \ref{tab:comp_other}, the proposed method outperforms previous approaches in task success rate. 
The overall performance of OpenVLA-ft (fine-tuned version on the Generated-3000 dataset in the RoboCasa dataset) remains considerably low, despite increasing the action token accuracy to over 97\% during training.
We hypothesize that OpenVLA, which has been pretrained with the OXE \cite{oxe_dataset} dataset focusing on real-world environments, may exhibit limited performance in simulation environments due to its specialization in real-world contexts.
Specifically, the OpenVLA-ft frequently struggles with the PnP variants. 
In comparison to tasks like ``CloseDrawer" or ``TurnOffSinkFaucet," PnP tasks require much more precise motions to accurately pick objects. 
This implies OpenVLA focuses on overall motion rather than fine-grained precision.

BC-Transformer, on the other hand, achieves a higher success rate than OpenVLA by focusing more on detailed motions. 
However, it often fails at seemingly simple tasks, such as opening and closing doors, except for ``CloseSingleDoor." 
This indicates that BC-Transformer struggles to differentiate between the detailed meanings of single/double doors or open/close actions.
In contrast, our DP-VLA achieves a higher success rate than BC-Transformer and maintains consistent performance across diverse tasks. 
The latent features extracted from the L-Sys2 provide a general understanding of task objectives, while the S-Sys1 module refines this with precise control over detailed motions. 
By leveraging the complementary strengths of prior approaches, our method demonstrates robust effectiveness across a wide variety of tasks.

\begin{table*}[!t]
\centering
\caption{Performance comparison of task success rates with other methods based on the RoboCasa evaluation protocol. OpenVLA-ft is a model fine-tuned on the RoboCasa dataset. BC-xfmr refers to a BC-Transformer model trained from scratch on the same dataset.}
\label{tab:comp_other}
\resizebox{\textwidth}{!}{
\begin{tabular}{ll|cc|c}
\toprule
\textbf{Category} & \textbf{Task} & OpenVLA-ft \cite{kim2024openvla}  & BC-xfmr \cite{robocasa2024} & \textbf{DP-VLA(ours)} \\
\midrule
\multirow{8}{*}{Pick and Place} & PnPCabToCounter & 0.00 & 0.18 & 0.10 \\
 & PnPCounterToCab & 0.00 & 0.28 & 0.32 \\
 & PnPCounterToMicrowave & 0.00 & 0.18 & 0.24 \\
 & PnPCounterToSink & 0.02 & 0.44 & 0.30 \\
 & PnPCounterToStove & 0.00 & 0.06 & 0.22 \\
 & PnPMicrowaveToCounter & 0.00 & 0.08 & 0.18 \\
 & PnPSinkToCounter & 0.00 & 0.42 & 0.56 \\
 & PnPStoveToCounter & 0.00 & 0.28 & 0.62 \\
\midrule
\multirow{4}{*}{Open/Close Doors} & OpenSingleDoor & 0.00 & 0.50 & 0.42 \\
 & OpenDoubleDoor & 0.00 & 0.48 & 0.80 \\
 & CloseDoubleDoor & 0.02 & 0.46 & 0.84 \\
 & CloseSingleDoor & 0.24 & 0.94 & 1.00 \\
\midrule
\multirow{2}{*}{Open/Close Drawers} & OpenDrawer & 0.00 & 0.74 & 0.66 \\
 & CloseDrawer & 0.72 & 0.96 & 1.00 \\
\midrule
\multirow{2}{*}{Twisting Knobs} & TurnOnStove & 0.06 & 0.46 & 0.64 \\
 & TurnOffStove & 0.06 & 0.24 & 0.16 \\
\midrule
\multirow{3}{*}{Turning Levers} & TurnOnSinkFaucet & 0.10 & 0.34 & 0.56 \\
 & TurnOffSinkFaucet & 0.56 & 0.72 & 0.72 \\
 & TurnSinkSpout & 0.36 & 0.96 & 0.90 \\
\midrule
\multirow{3}{*}{Pressing Buttons} & CoffeePressButton & 0.00 & 0.74 & 0.86 \\
 & TurnOnMicrowave & 0.08 & 0.90 & 0.84 \\
 & TurnOffMicrowave & 0.02 & 0.60 & 0.86 \\
\midrule
\multirow{2}{*}{Insertion} & CoffeeServeMug & 0.12 & 0.34 & 0.64 \\
 & CoffeeSetupMug & 0.00 & 0.12 & 0.30 \\
\midrule
\multicolumn{2}{c|}{\textbf{Avg. Task Success Rate}} & 0.098 & 0.476 & \textbf{0.573} \\
\bottomrule
\end{tabular}}
\end{table*}

\subsubsection{Inference Speed}

Another key advantage of our approach is the high inference speed it achieves. By primarily executing the S-Sys1 and only occasionally computing the L-Sys2, our method overcomes the slow inference limitation of OpenVLA. 
As shown in Table \ref{tab:inference_time}, our DP-VLA achieves an impressive inference time of 0.03 seconds, closely matching the speed of BC-Transformer, while demonstrating 20.4\% better performance.
Notably, it is significantly faster than OpenVLA, demonstrating that our method accelerates inference while simultaneously improving task success rates.

\begin{table*}[h]
\centering
\caption{Performance comparison of average inference time with other methods. The average time refers to the mean inference time measured from 1st to 50th frame of an episode. BC-xfmr encodes the clip language encoder only at the initial frame of the episode, and similarly, our DP-VLA performs L-Sys2 encoding only at the initial frame.}
\label{tab:inference_time}
\begin{tabular}{lcc|c}
\toprule
 & OpenVLA-ft \cite{kim2024openvla} & BC-xfmr \cite{robocasa2024} & \textbf{DP-VLA (ours)}\\
\midrule
\textbf{Avg. Inference Time} (sec) & 0.253 & \textbf{0.022} & 0.030 \\
\bottomrule
\end{tabular}
\end{table*}

\subsection{Ablation Study}
Additionally, our method was evaluated through two ablation studies: the first investigates the selection of latent feature types, while the second examines the necessity of tuning the L-Sys2 model.

\subsubsection{Comparisons across Latent Feature Types}

As an ablation study, we analyzed which representation from OpenVLA works best for dual-processing. 
In our design, selecting the appropriate latent features from OpenVLA to transfer to S-Sys1 is crucial. 
These features must include an understanding of the given commands from text inputs as well as information about the rough trajectory needed to accomplish the task. 
To verify this, we evaluated four types of latent features extracted from OpenVLA. 
For this experiment, we trained each S-Sys1 model on the Generated-3000 dataset with a batch size of 128 for 200,000 iterations, using latent features from the OpenVLA-ft model (fine-tuned).
\begin{enumerate}
    \item \textbf{Mean feature of input text prompt}: an averaged feature of the latent features from the input text sequence.
    \item \textbf{Intermediate feature at the end of input text prompt}: a latent feature corresponding to the last token in the input text sequence.
    \item \textbf{Intermediate feature at the start of action decoding}: a latent feature corresponding to the x-axis movement represented by the first output token.
    \item \textbf{Intermediate feature at the end of action decoding}: a latent feature preceding the gripper motion prediction.
\end{enumerate}

\vspace{-15pt}
\begin{table*}[h]
\centering
\caption{Comparison across feature types: mean feature of input text prompt, feature at the end of input text prompt, feature at the start of action decoding, and feature at the end of action decoding.} 
\label{tab:latent_features}
\resizebox{\textwidth}{!}{
\begin{tabular}{l|c|c|c|c}
\toprule
 & \multicolumn{2}{c|}{\textbf{Prefill Stage}} & \multicolumn{2}{c}{\textbf{Decoding Stage}} \\
\cmidrule(r){2-3} \cmidrule(r){4-5}
 & Mean-of-Text & End-of-Text & Start-of-Action & End-of-Action \\
\midrule
\textbf{Avg. Task Success Rate} & 0.533 & 0.490 & \textbf{0.543} & 0.520 \\
\bottomrule
\end{tabular}}
\end{table*}

As shown in Table \ref{tab:latent_features}, the latent features from the decoding stage slightly outperform those from the prefill stage. 
In typical VLMs, the next token is predicted during both stages, but the predictions from the prefill stage have only an indirect effect on the final output. 
In contrast, predictions made during the decoding stage are critical, as they directly influence the final outputs. 
As a result, the prefill stage features are generally unused, making the output-stage features more practically relevant. 
This distinction suggests that the decoding stage features are likely to contain more task-relevant, motion-related information.


\subsubsection{Analysis about the Effectiveness of Finetuning}

We compared the contributions of the pretrained and finetuned OpenVLA as L-Sys2. 
It is well known that the pretrained OpenVLA captures general knowledge about robot manipulation. 
In this analysis, we compared the performance of general knowledge (from the pretrained model) and task-specific knowledge (from the finetuned model) in their roles as L-Sys2.
For this experiment, we trained each S-Sys1 model on the Generated-3000 dataset with a batch size of 128 for 300,000 iterations, using latent features of the End-of-Text type.

Interestingly, DP-VLA utilizing latent features from the pre-trained OpenVLA outperformed the one based on features from the fine-tuned OpenVLA as described in Table \ref{tab:pt_vs_ft}.
This observation indicates that fine-tuning may have diminished OpenVLA’s general understanding of tasks, potentially sacrificing some of its broader task knowledge. 
Thus, fine-tuning only the S-Sys1 is sufficient, highlighting the efficiency of the training process.

\vspace{-5pt}
\begin{table*}[h]
\centering
\caption{Performance comparison based on latent features extracted from the pre-trained OpenVLA (OpenVLA-pt) and the fine-tuned OpenVLA (OpenVLA-ft)}
\label{tab:pt_vs_ft}
\begin{tabular}{l|c|c}
\toprule
 & \multicolumn{2}{c}{\textbf{Large System 2 Model}} \\\cmidrule(r){2-3}
 &  OpenVLA-pt & OpenVLA-ft \\
\midrule
\textbf{Average Task Success Rate} & \textbf{0.556} & 0.512 \\
\bottomrule
\end{tabular}
\end{table*}
\vspace{-5pt}

\section{Future Work}
\label{sec:future_work}
In this paper, we divide the entire VLA process into two sub-processes: the S-Sys1 and L-Sys2, inspired by the dual-process theory of human cognition. However, we believe that this approach can be further optimized for VLA tasks through a more precise, adaptive design. 
A key consideration is designing a dynamic scheduling mechanism for L-Sys2, where its execution frequency adapts to task requirements by introducing a supervisory module that determines when L-Sys2 is necessary, allowing it to skip complex reasoning during simpler tasks and adjust its involvement based on task or environment complexity.
Additionally, the L-Sys2 can be decomposed into multiple levels, each operating at different frequencies depending on the level of reasoning or decision-making required. 
By refining the design and scheduling of these processes, VLA models can achieve greater efficiency and faster inference times.
We also plan to further our research on robotic manipulation by investigating advanced architectures and learning strategies for DP-VLA. 
Furthermore, we will experiment with various VLMs to leverage latent features that can further extend the model's capabilities. For example, these features can be utilized for long-horizon task planning or improving contextual understanding through Chain of Thought reasoning.
Building on these improvements, we aim to extend this work by developing a more precise and adaptive framework, optimizing the balance between computational efficiency and task performance.

\section{Conclusion}
\label{sec:conclusion}


In this work, we have introduced Dual Process VLA, inspired by the dual-process theory from human cognitive psychology. This approach involves partitioning the VLA process into two distinct modules: L-Sys2, which processes complex reasoning and high-level decision-making tasks less frequently, and S-Sys1, which manages routine, frequent tasks such as sensory processing, intermediate action prediction, and motor control. 
By scheduling L-Sys2 operations less frequently while S-Sys1 operates continuously, we achieve both computational efficiency and smooth, natural robotic actions.
Our experimental results demonstrate that the proposed dual-process VLA framework effectively reduces redundant computations and overcomes the limitations of prior VLA models, such as high computational burden and discontinuous actions. This hierarchical separation provides a flexible, efficient approach for real-time action prediction, enhancing both the responsiveness and overall performance of robots in diverse environments. 
We believe that Dual Process VLA represents a significant step toward more efficient and natural robotic behavior in complex tasks involving vision, language, and action. This approach has the potential to contribute to a future where humans and robots coexist seamlessly.




\clearpage
\acknowledgments{This work was partly supported by Electronics and Telecommunications Research Institute (ETRI) grant funded by the Korean government foundation (24ZB1200, Research of Human-centered Autonomous Intelligence System Original Technology).}


\bibliography{Han_CoRL24w}  

\end{document}